\documentclass[letterpaper, 10 pt, journal, twoside]{IEEEtran}  

\IEEEoverridecommandlockouts                              %

\usepackage{graphics} %
\usepackage{epsfig} %
\usepackage{times} %
\usepackage{amsmath} %
\usepackage{amssymb}  %
\usepackage{amsfonts}
\usepackage{hyperref} %
\usepackage{xcolor} %
\usepackage{arydshln} %
\usepackage{color} 
\usepackage{booktabs}
\usepackage{hyperref}
\usepackage{float}
\usepackage{multirow}
\usepackage{pifont}
\usepackage{bm}
\usepackage{tikz}
\usepackage{tabularx}
\usepackage{tikz-3dplot}
\usepackage{multirow}
\usepackage{mathtools}%
\usepackage{graphicx}
\usepackage{units}
\usepackage{balance}
\usepackage{cite}
\usepackage{pgfplots}
\usepackage{pgfplotstable}
\usetikzlibrary{bending, external, plotmarks, shapes, pgfplots.statistics, pgfplots.colorbrewer, positioning}
\usepgfplotslibrary{fillbetween}
\pgfplotsset{compat=1.16} 
\usepackage{xfrac}

\newcommand{\bfr}[0]{\ensuremath{\mathcal{B}}} %
\newcommand{\ffr}[0]{\ensuremath{\mathcal{F}}} %

\newcommand{\rotateRPY}[3]%
{   \pgfmathsetmacro{\rollangle}{#1}
    \pgfmathsetmacro{\pitchangle}{#2}
    \pgfmathsetmacro{\yawangle}{#3}

    \pgfmathsetmacro{\newxx}{cos(\yawangle)*cos(\pitchangle)}
    \pgfmathsetmacro{\newxy}{sin(\yawangle)*cos(\pitchangle)}
    \pgfmathsetmacro{\newxz}{-sin(\pitchangle)}
    \path (\newxx,\newxy,\newxz);
    \pgfgetlastxy{\nxx}{\nxy};

    \pgfmathsetmacro{\newyx}{cos(\yawangle)*sin(\pitchangle)*sin(\rollangle)-sin(\yawangle)*cos(\rollangle)}
    \pgfmathsetmacro{\newyy}{sin(\yawangle)*sin(\pitchangle)*sin(\rollangle)+ cos(\yawangle)*cos(\rollangle)}
    \pgfmathsetmacro{\newyz}{cos(\pitchangle)*sin(\rollangle)}
    \path (\newyx,\newyy,\newyz);
    \pgfgetlastxy{\nyx}{\nyy};

    \pgfmathsetmacro{\newzx}{cos(\yawangle)*sin(\pitchangle)*cos(\rollangle)+ sin(\yawangle)*sin(\rollangle)}
    \pgfmathsetmacro{\newzy}{sin(\yawangle)*sin(\pitchangle)*cos(\rollangle)-cos(\yawangle)*sin(\rollangle)}
    \pgfmathsetmacro{\newzz}{cos(\pitchangle)*cos(\rollangle)}
    \path (\newzx,\newzy,\newzz);
    \pgfgetlastxy{\nzx}{\nzy};
}

\newcolumntype{C}{>{\centering\arraybackslash}X}
\newcolumntype{x}[1]{>{\centering\let\newline\\\arraybackslash\hspace{0pt}}p{#1}}
\definecolor{matlab1}{rgb}{0.00000,0.44700,0.74100}
\definecolor{matlab2}{rgb}{0.85000,0.32500,0.09800}
\definecolor{matlab3}{rgb}{0.92900,0.69400,0.12500}
\definecolor{matlab4}{rgb}{0.49400,0.18400,0.55600}
\definecolor{matlab5}{rgb}{0.4660, 0.6740, 0.1880}
\definecolor{matlab6}{rgb}{0.3010, 0.7450, 0.9330}
\definecolor{matlab7}{rgb}{0.6350, 0.0780, 0.1840}
\definecolor{matlab8}{rgb}{0.8, 0.8, 0}
\definecolor{matlab9}{rgb}{0.6, 0.6, 0.6}
\definecolor{verylightgray}{rgb}{0.98,0.98,0.98}

\newcommand{\rebuttal}[1]{\textcolor{black}{#1}}
\pgfdeclarelayer{bg}

\title{
Robotics meets Fluid Dynamics: A Characterization of the Induced Airflow below a Quadrotor as a Turbulent Jet
}

\author{Leonard Bauersfeld,$^{1}$  Koen Muller,$^{2}$ Dominic Ziegler,$^{2}$ Filippo Coletti,$^{2}$ Davide Scaramuzza$^{1}$ \\ 
\thanks{Manuscript received: Aug. 29, 2024; Revised: Oct 31, 2024; Accepted: Nov 28, 2024.
This paper was recommended for publication by Editor Giuseppe Loianno upon evaluation of the Associate Editor and Reviewers' comments.}
\thanks{This work was supported by the European Union’s Horizon Europe Research and Innovation Programme under grant agreement No. 101120732 (AUTOASSESS), the European Research Council (ERC) under grant agreement No. 864042 (AGILEFLIGHT), the Swiss National Science Foundation under project No. 200021\_212065, and the Dutch Organisation for Scientific Research
(NWO) Rubicon grant project No. 019.233EN.018.}
\thanks{$^1$ L. Bauersfeld and D. Scaramuzza are with the Robotics and Perception Group, University of Zurich, Switzerland {\tt\footnotesize bauersfeld@ifi.uzh.ch}}
\thanks{$^2$ K. Muller, D. Ziegler, and F. Coletti are with the Institute of Fluid Dynamics, ETH Zurich, Switzerland}
\thanks{Digital Object Identifier (DOI): see top of this page.}
}

\begin{document}

\makeatletter
\maketitle

\begin{abstract}
The widespread adoption of quadrotors for diverse applications, from agriculture to public safety, necessitates an understanding of the aerodynamic disturbances they create. 
This paper introduces a computationally lightweight model for estimating the time-averaged magnitude of the induced flow below quadrotors in hover. 
Unlike related approaches that rely on expensive 
computational fluid dynamics (CFD) simulations or drone specific time-consuming empirical measurements, our method leverages classical theory from turbulent flows.
By analyzing over 16 hours of flight data from drones of varying sizes within a large motion capture system, we show for the first time that the combined flow from all drone propellers is well-approximated by a turbulent jet after 2.5 drone-diameters below the vehicle.
Using a novel normalization and scaling, we experimentally identify model parameters that describe a unified mean velocity field below differently sized quadrotors.
The model, which requires only the drone's mass, propeller size, and drone size for calculations, accurately describes the far-field airflow over a long-range in a very large volume which is impractical to simulate using  CFD.
Our model offers a practical tool for ensuring safer operations near humans, optimizing sensor placements and drone control in multi-agent scenarios.
We demonstrate the latter by designing a controller that compensates for the downwash of another drone, leading to a four times lower altitude deviation when passing below.
\\ Video: \small{\url{https://youtu.be/-erfmxWTzPs}}
\end{abstract}

\vspace*{-6pt}
\section{Introduction}

\IEEEPARstart{I}{n} recent years, quadrotors have gained popularity for a wide variety of tasks in academia~\cite{kaufmann23champion, 2014:DAndrea, 2017:Karydis} as well as in industry where companies develop drones for filming~\cite{djiweb}, mapping~\cite{flyabilityweb}, inspection~\cite{skydioweb}, and public safety~\cite{parrotweb}. 
Shared across these diverse applications is the need to understand and characterize the strong downwash generated by the vehicle's propellers as this enables informed decisions on the allowed proximity of a quadrotor to an object or person, ultimately leading to safer autonomous drones.

In mapping and inspection tasks this helps predicting when aerodynamic interactions with close-by structures such as bridges, powerlines~\cite{AerialCore}, or ships~\cite{AutoAssess} occur. 
When drones are operated for agricultural purposes, it is critical to know how far the aerodynamic disturbances caused by the drone extend for plant spraying and protection~\cite{Chang2023}. 
When deployed for filming and in public spaces, quadrotors are often operated in the vicinity of people where minimizing the presence of intrusive flows in the scene is important. 
In scenarios where multiple drones are operated together, a flow model can be used to improve planning such that individual vehicles dynamically avoid each other's downwash and show an improved response to external disturbances~\cite{Simon2023}. 
Finally, a better understanding of the flow can be important for sensor and scientific instrumentation placement~\cite{Villa2016,Crazzolara2019,McKinney2019,Thielicke2021,Ghirardelli2023}. This paper presents a computationally lightweight approach to model the time-averaged magnitude of the induced flow below a quadrotor at hover. 

\begin{figure}[t!]
    \centering
    \includegraphics{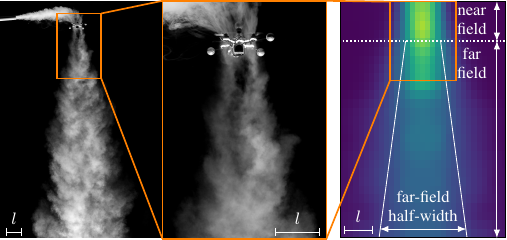}
    \vspace*{-8pt}
    \caption{\rebuttal{Smoke visualization of the flow (left) and corresponding measured velocity field (right) of the \emph{Kolibri}} drone at hover. The individual propellers flows are separated close to the drone. After 2.5 motor-to-motor distances $l$ the individual flows merge into one turbulent jet (far-field) for which the velocity field and half-width (distance where the velocity is half the centerline velocity) can be calculated with our proposed method.}
    \label{fig:fig1}
    \vspace*{-18pt}
\end{figure}

Detailed computation of the induced flow around a drone is a challenging problem as it requires expensive and time-consuming computational fluid dynamics (CFD) simulations. 
To limit the computational demand, CFD simulations typically focus on millisecond time scales and simulate the airflow only close to the vehicle~\cite{paz2021assesmentcfd, ventura2018high, luo2015novel}. 
Such a simulation-based methodology is suitable to assess the drone flight performance, but not efficient in analyzing the far-field which extends over many meters away from the drone where one needs to rely on turbulence modeling \cite{McKinney2019, Ghirardelli2023}. 
Additionally, slow CFD simulations are not usable in a real-time planner or controller to dynamically coordinate a fleet of quadrotors, such that the vehicles minimize interference with each other.

Orthogonal to the simulation approach, fluid dynamics research has a century-old history of heavily relying on empirical measurement. 
However, experimentally characterizing each drone with thousands of measurements in the entire 3D space around it is a tedious process requiring a large, wind-free measurement domain. 
Additionally, a proper measurement technique needs to be selected. 
For example particle-image-velocimetry (PIV)~\cite{Adrian2011} is well suited, but generally presents  challenges in large-scale real-world deployments~\cite{Kuhn2010,Jux2018}.
\vspace*{-5pt}
\subsection*{Contribution}
\vspace*{-1pt}

\rebuttal{Our main contribution is} a unified, computationally lightweight model to calculate the mean velocity of the induced flow below a drone at hover. Or, put differently, we answer the question: `How much wind does a drone generate when hovering or flying slowly in a near-hover state?' The unified model is inspired by decades of well-established research on turbulent jets~\cite{Pope2000}. 
The advance in this work is enabled by joining methodologies from fluid dynamics and robotics research. 
Recording over \unit[16]{h} of drone flight data with six differently sized drones ranging from \unit[230]{g} to \unit[6.3]{kg} we perform pointwise flow measurements in a very large motion capture system~\cite{Mueller2012}. %
Through the use of appropriate normalization and the introduction of a characteristic drone length-scale the model is unified and can be applied to quadrotors of different mass and size. 

Our methodology is computationally lightweight as it does not rely on CFD simulations. Instead, it uses available closed-form analytic flow solutions. All calculations only require knowledge of the vehicle's mass, the vehicle's dimensions, and the propeller size. We summarize our model in a pen-and-paper algorithm that calculates the downwash velocity field, making it easy to apply our findings to tasks in other domains, such as agriculture and public safety.

\rebuttal{Additionally, to} demonstrate the \rebuttal{applicability} of our simple method to real-world robotics applications, we integrate it into a controller that automatically compensates for downwash when passing below another drone, yielding a four times smaller altitude tracking error.

\vspace*{-4pt}
\section{Related Work}
\vspace*{-1pt}
For control and simulation tasks the primary goal of a model is to predict the motion of the vehicle given a certain actuation, that is, predict the forces and torques acting on its body. 
The most widespread models are rotor-based quadratic thrust and drag models~\cite{mahony2012multirotor, furrer2016rotors, shah2018airsim} and more advanced models that are  based on blade-element-momentum (BEM) theory which calculates the lift and drag at a
propeller based on airfoil theory~\cite{1995:Prouty, 2021:Bauersfeld, khan2013toward, gill2017propeller, gill2019computationally, hoffmann2007quadrotor, huang2009aerodynamics}.
However, these approaches focus on estimating the aerodynamic forces and torques of the individual rotors but do not compute the combined total induced volumetric flow.

In contrast to the above models, computational fluid dynamics simulations do calculate the three-dimensional flow around the vehicle.
Such approaches are typically used to determine and improve the efficiency of the design of a single propeller~\cite{ragni2011non, westmoreland2008modeling}, but have also been applied to simulate entire quadrotors~\cite{luo2015novel, ventura2018high}, and larger hexacopters~\cite{Zheng2018,McKinney2019}. 
While such simulations achieve results that are highly accurate and manage to capture real-world effects%
~\cite{yasuda2013numerical, yoon2017computationalmodeling}, they require large amounts of computation. 
Additionally, they do not easily generalize to other vehicles, in principle requiring a new analysis for each and every drone design.

In contrast to a purely simulation-based approach, a few works perform real-world experiments. 
For example,~\cite{Villa2016} performed flow probe measurements on a hexacopter that seem to find overall agreement with later CFD work \cite{McKinney2019}. 
In~\cite{paetzhold2023flight} the authors present experimental data to understand the influence of the quadrotor on atmospheric temperature and pressure measurements. 
In~\cite{liu2022extraction} the authors focus on the development of a Schlieren-photography method to qualitatively visualize the flow around drones. 
In \cite{Crazzolara2019} the authors performed outdoor smoke visualization, and in \cite{Otsuka2017} the authors applied PIV to smoke visualizations to estimate the velocity field around a quadrotor model in ground effect.

Only a few other studies have applied PIV to drones.
For example, in~\cite{Shukla2018} the interaction between propellers under different configurations has been studied, and in~\cite{Czyz2020} the near field flow of a quadrotor model in forward flight is analyzed using wind tunnel experiments. 
Most recently ~\cite{Chang2023} performed PIV in a different application and recovered the spreading rate of spray droplets in the drone's downwash.
Apart from~\cite{Crazzolara2019,paetzhold2023flight}, all of these works performed measurements in tethered flight, presenting actual limitations in recovering the true free flight physics. To the authors' best knowledge, no studies have looked into a comprehensive analysis of the flow below drones in hover while focusing on a scaling analysis, and on closed-form solutions from turbulent flows.

\vspace*{-4pt}
\section{Theory}
\vspace*{-3pt}
This section gives an overview of the relevant notation, overall drone thrust and propeller downwash in hovering flight, and introduces the concept of a turbulent jet flow.
Finally, we introduce several normalizations enabling the analysis to generalize across a wide variety of drones.

\begin{figure}[t!]
    \centering
    \includegraphics{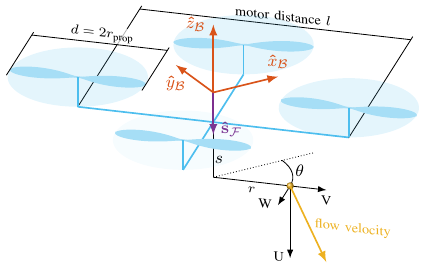}
    \vspace*{-10pt}
    \caption{Diagram for the drone and the flow coordinate system and geometry. 
    The \textcolor{matlab2}{bodyframe} $\bfr$ of a \textcolor{matlab6}{quadrotor} is such that the x-axis faces forward, and the z-axis upwards in thrust direction. 
    The \textcolor{matlab4}{flow coordinate system} is oriented such that its longitudinal axis $\mathbf{\hat{s}_\ffr}$ is aligned with flow direction (i.e., points in negative $\mathbf{\hat{z}_\bfr}$ direction). 
    The \textcolor{matlab3}{flow} is described in cylindrical coordinates $[s~~r~~\theta]^\top$ by its longitudinal (axial), radial, and azimuthal velocity components $U$, $V$ and $W$, respectively.
    The vehicle's propeller radius $r_\text{prop}$, diameter $d$ and motor-distance $l$ are defined as depicted.
    }
    \label{fig:quadrotor}
    \vspace*{-13pt}
\end{figure}

\vspace*{-6pt}
\subsection{Coordinate System and Notation}
\vspace*{-1pt}

Figure~\ref{fig:quadrotor} shows the quadrotor-centered bodyframe coordinate system $\bfr$, with vector $\mathbf{x}_\bfr=[x~~y~~z]^\top$. 
In the bodyframe the $\mathbf{\hat{x}_\bfr}$-axis points forward, the $\mathbf{\hat{y}_\bfr}$-axis points to the left, and the $\mathbf{\hat{z}_\bfr}$-axis points upwards. 
Since we only consider the quadrotor at hover, the $\mathbf{\hat{z}_\bfr}$-axis is opposing the gravity vector at all times and no rotation due to the roll and pitch of the vehicle needs to be considered. 
We also introduce a flow coordinate system $\ffr$ with $\mathbf{x}_\ffr = [s~~r~~\theta]^\top$ in cylindrical coordinates along the downward flow direction $\mathbf{\hat{s}_\ffr}$. 
The longitudinal coordinate $s$ is aligned with the downwash  and the radial coordinate $r$ points outwards as shown in Fig.~\ref{fig:quadrotor}. 
In this coordinate system the velocity vector $\mathbf{U}$ with components $U, V, W$ is defined as shown in Fig.~\ref{fig:quadrotor}. 

\vspace*{-6pt}
\subsection{Propeller-Induced Flow}
\vspace*{-1pt}
In a steady-state hover, the multicopter must produce enough thrust to support its weight. 
This means that, considering a quadrotor, the \rebuttal{$N_P=4$} propellers must produce a thrust force that is equal to $mg$ where $m$ is the mass and $g$ is the gravitational acceleration. 

A propeller produces lift force by accelerating air downwards. 
Assuming that the air above the propeller is at rest and that the air is accelerated in a virtual flow tube around the propeller a momentum balance across the propeller yields the induced-velocity $U_{\rm{H}}$  at hover~\cite{1995:Prouty}:
\begin{equation}
    U_{\rm{H}} = \sqrt{\frac{T_{\rm{H}}}{2 \rho A_\text{prop}}} = \sqrt{\frac{\vphantom{T}m g}{2 \rho \pi r_\text{prop}^2 N_P}}\ ,
    \label{eq:v_ind_h}
\end{equation} %
where $T_{\rm{H}}$ is the hover thrust and $\rho$ is the air density. 
For the definition above, we assume that the flow tube has the same width as the propeller diameter, e.g. the induced velocity is calculated directly below the propeller. 
This model is an oversimplification and, for example, does not capture the true velocity profile of the flow below the propeller. 
However, as the model represents an overall momentum balance it can be interpreted as an average velocity across the flow tube close to the propeller~\cite{1995:Prouty,hoffmann2007quadrotor,2021:Bauersfeld}. 

The air-density $\rho$ is computed according to the ideal gas law~\cite{ICAO1954} to account for the local temperature $\vartheta$ and absolute pressure $p$ as
\begin{equation}
    \rho = p M/(R \vartheta)
\end{equation}
where $M=\unit[28.966]{g/mol}$ the molar mass of dry air and ${R=\unit[8.3144]{J/(K\cdot mol)}}$ is the universal gas constant.

\vspace*{-6pt}
\subsection{Turbulent Jet Flow}
\vspace*{-1pt}
The key idea of our simple quadrotor aerodynamic model is to approximate the combined rotor-induced flow as a turbulent jet~\cite{Pope2000}.
Because the airflow is turbulent, we do not consider the instantaneous flow values, instead, we focus on time-averaged flow variables denoted by the averaging brackets $\langle \cdot \rangle$.

For a turbulent jet, the mean longitudinal flow velocity $\langle U \rangle$ along the radial and flow-direction coordinates $r,s$ is given by the similarity profile~\cite{Pope2000}:
\begin{equation}
    \langle U \rangle(\xi) =\frac{ U_\text{C}(s) }{\left(1 + \left( \sqrt{2}-1 \right) \xi^2 \right)^2}\; .
    \label{eq:radialprofile}
\end{equation} %
Here $U_\text{C}$  describes the centerline velocity at $r=0$ as a function of the distance $s$ away from its exit nozzle, here below the rotor plane.  For a jet with half-width $r_{1/2}$, the rescaled radial position $\xi$ is given by:
\begin{equation}
    \xi = r\,/\,r_{\raisebox{-1.5pt}{\scriptsize $1/2$}}(s)\ .
    \label{eq:xi}
\end{equation} %
The jet centerline velocity is known to scale inversely proportional to the distance from its exit. It is given by
\begin{equation}
     U_\text{C} (s) =  U_\text{J}  \frac{B\,d}{s - s_0}\ ,
    \label{eq:centerline}
\end{equation} %
where $U_\text{J}$ is initial jet velocity, $d$ the jet exit diameter, $B$ an empirical constant, and $s_0$ is the flow development-length.
In addition, the jet half-width spreads linearly as
\begin{equation}
    r_{1/2}(s)=S(s-s_0)\ ,
    \label{eq:spread}
\end{equation} 
where $S$ is the spreading rate that relates to the jet opening angle $\theta=2\arctan{(S)}$.

A key characteristic of the turbulent jet is that the spreading angle is commonly around \unit[12]{deg} and is weakly dependent on the flow Reynolds number $Re_{\rm{d}}=\rho U d / \mu$ that describes the relative importance between inertial and viscous forces~\cite{Pope2000}.  This makes the model applicable to a wide range of drone flow and geometry, and independent of the air-viscosity $\mu$.
We remark that we do not consider propeller swirl in the analysis as the rotors counter-rotate, and we assume that all other flow components are entrained (drawn/sucked into the flow from the side) downstream.

\vspace*{-6pt}
\subsection{Normalization}
\vspace*{-1pt}
\label{sec:normalization}
To develop a generalized model of the airflow around different drones, the most important physical properties of those drones must be taken into account. 
We achieve this by introducing a velocity and a spatial normalization based on the theoretical considerations described previously. 

\subsubsection*{Velocity Normalization} From \eqref{eq:radialprofile} and \eqref{eq:centerline} we observe that the overall scaling of the jet velocity is given by the jet-exit velocity $U_\text{J}$. 
The jet exit velocity describes how fast a jet exits a nozzle and captures how much momentum the flow carries. 
While a quadrotor does not have a nozzle, the induced velocity $U_{\rm{H}}$ of eq.~\eqref{eq:v_ind_h} relates to the same physical quantity.
Denoting a normalized quantity with~$\sim$, the normalized velocity $\tilde{U}$ and normalized centerline velocity $\tilde{U}_C$ are given by:
\begin{equation}
     \langle\tilde{U} \rangle (\xi) = \langle U\rangle (\xi)/ U_{\rm{H}} \quad\text{and} \quad\tilde{U}_{\text{C}}(s)= U_\text{C}(s)/ U_{\rm{H}}\;.
    \label{eq:vel_normlaization}
\end{equation}%

\subsubsection*{Length-Scale Normalization} Similar to the velocity normalization, a spatial normalization parameter is needed to be able to compare drones of different sizes.
We propose the motor distance $l$ as the length-scale parameter as this distance is found to be the closest equivalent to a nozzle's diameter in a jet flow. 
Using this, we define the normalized distance to the rotor plane $\tilde{s}$, the normalized radial distance $\tilde{r}$ and the normalized half-width $\tilde{r}_{1/2}$ as:
\begin{equation}
    \tilde{s} = s/l,\quad\tilde{r}=r/l,\quad\text{and}\quad\tilde{r}_{1/2}(s)=r_{1/2}(s)/l\;.
\end{equation}

\section{Flight Experiments}

\subsection{Experimental Setup}

\begin{figure}[t!]
    \centering
    \includegraphics{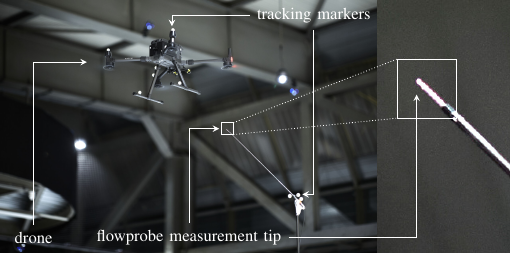}
    \vspace*{-8pt}
    \caption{The \emph{DJI Matrice 300} drone hovers in proximity to the flow probe. The inset shows a close-up view of the Testo hot ball probe. The flow probe measures the speed of the airflow.}
    \label{fig:experimental_setup}
    \vspace*{-15pt}
\end{figure}

Measurements of the flow around a drone require a large indoor space to avoid  air recirculation while preventing external disturbances such as wind. 
\rebuttal{For this study a large industrial hall with a 36-camera Vicon motion-capture system ($\unit[25\times 25\times 7]{m}$ tracking volume) is used to record the drone's position with millimeter accuracy.}

The flow is measured with an omnidirectional hot ball flow probe (Testo 440) shown in Fig.~\ref{fig:experimental_setup}. 
This type of commercial, off-the-shelf anemometer is indifferent to the flow direction and presents a sufficiently high measurement accuracy of {$\pm \unit[0.03]{m/s} \pm 5\%$} (up to \unit[20]{m/s}) at a sampling rate of \unit[1]{s}. 
This is favorable as we are interested in the induced mean magnitude and not the fluctuating velocity components of vortices in the turbulent flow. 

\vspace*{-6pt}
\subsection{Quadrotors}
For experiments, six different quadrotors are used whose properties are summarized in Tab.~\ref{tab:drones}. 
\emph{Kolibri}, \emph{Offboard~1} and \emph{Offboard~2} are research drones that can autonomously fly missions~\cite{foehn2022agilicious} in the motion-capture system. 
Both \emph{Offboard} drones share the same frame, however, drone 2 has double the mass compared to drone 1. 
\rebuttal{To demonstrate the generality of the proposed jet flow model we also use a \emph{DJI Matrice 300} drone as a much larger and heavier vehicle.}
The aforementioned drones have an uncanted propeller arrangement, meaning that their axis of rotation is parallel to the $z_\bfr$-axis.
We additionally use two commercial drones, respectively a \emph{DJI Mavic 3E} (enterprise) and a \emph{Flyability Elios 3}, with canted propeller configurations to better understand the limitations of our model. 
The commercial \emph{DJI} and \emph{Flyability} drones must be flown manually throughout the entire experiment.

\vspace*{-6pt}
\subsection{Data Collection}
To collect the experimental data, each vehicle is flown in a grid-like pattern where the drone approaches a point, steadily hovers for \unit[5]{s} to allow the flow to (re-)develop, and then slowly translates to the next point in the three-dimensional flight path. 
For the autonomous drones, we sample a coarse xy-grid spanning ${10\times 10}$ length-scales (resolution 0.66~$l$) and a fine xy-grid spanning ${3\times 3}$ length-scales (resolution 0.33~$l$). 
In $z$ we cover a range of up to 3m above the probe. 
For the \emph{Mavic 3E} and the \emph{Elios 3} the pilot sampled a similar pattern to the autonomous drone. 
For the Matrice 300 the pilot covered an area amounting to $\unit[5\times 6\times 4.5]{m}$.

\rebuttal{We filter out all data points where the vehicle moves with a speed greater than \unit[0.1]{m/s} to reduce the influence of transient effects.}
Despite being in a large indoor space, the anemometer registers a small ambient flow in the range of \unit[0.06]{m/s} to \unit[0.12]{m/s}. 
As this flow is already observed before the drone takes off, it is not primarily caused by recirculation effects but related to the testing facility. 
The background level is about 2 orders of magnitudes below the induced velocities and we correct for this static offset by subtracting the ambient airflow speed from the measurements. 
After filtering we obtain around $\unit[10000]{s}$ of flight data (position and corresponding anemometer measurement) for each of the vehicles. 
For further processing, this scattered data is binned into gridded data as the median of all measurements within a grid cell. 

\begin{table}[t!]
    \centering
    \caption{Overview of the quadrotors.}
    \vspace*{-6pt}
    \setlength{\tabcolsep}{4pt}
    \begin{tabularx}{1\linewidth}{l|cCCCC}
        \toprule
         &  Mass & Propeller Cant & Propeller Diameter & Motor Distance & Induced Velocity \\
         \midrule
         \emph{Kolibri} &     \unit[0.230]{kg} & uncanted & \unit[7.37]{cm}  & \unit[11.8]{cm} & \unit[7.41]{m/s} \\
         \emph{Offboard 1} &  \unit[0.572]{kg} & uncanted & \unit[12.95]{cm} & \unit[26.6]{cm} & \unit[6.66]{m/s} \\
         \emph{Offboard 2} &  \unit[1.207]{kg} & uncanted & \unit[12.95]{cm} & \unit[26.6]{cm} & \unit[9.66]{m/s} \\
         \emph{Matrice 300} & \unit[6.300]{kg} & uncanted & \unit[53.34]{cm} & \unit[89.4]{cm} & \unit[5.36]{m/s} \\
         \midrule  
         \emph{Mavic 3E} & \unit[0.958]{kg} & inward & \unit[23.88]{cm} & \unit[38.5]{cm} & \unit[4.67]{m/s} \\
         \emph{Elios 3} &     \unit[2.398]{kg} & outward & \unit[12.70]{cm} & \unit[27.5]{cm} & \unit[13.89]{m/s}\\
         \bottomrule
    \end{tabularx}
    \label{tab:drones}
    \vspace*{-15pt} 
\end{table}

\vspace*{-4pt}
\section{Results: Near Field}
In a first analysis, we focus on the flow directly below the drone. In this region the flow contributions of the four individual propellers are separated and have not yet fully merged into a combined drone downwash.
The goal is to analyse how the individual propeller flows eventually merge into one turbulent jet for the different quadrotors. 

\begin{figure}[t!]
    \centering
    \includegraphics{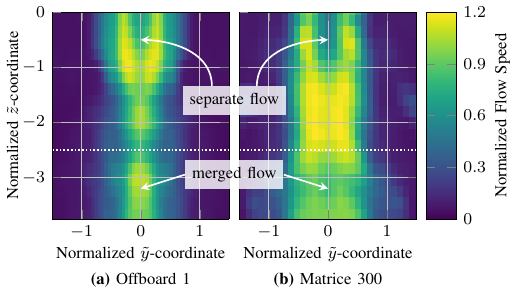}
    \vspace*{-12pt}
    \caption{Visualization of the near field of \textbf{(a)} the \emph{Offboard 1} drone and \textbf{(b)} the \emph{Matrice 300}.
    The length and velocity scales are normalized. The influence of the individual  propellers is clearly visible at $z=0$ and diminishes at a normalized distance of about 2.5 as indicated by the dotted white line.}
    \label{fig:nearfield}
    \vspace*{-6pt}
\end{figure}

Figure~\ref{fig:nearfield} exemplarily visualizes the flow field for a $yz_\bfr$-slice from the measurement data. 
The quadrotor is located at the `top' of the plot (e.g. at $\tilde{z}$=0) and one can clearly see the two separate flows  induced by the left and the right propellers.
The \emph{Offboard 1} and \emph{Matrice 300} drone show similar flow patterns as we normalize the axes with the drones' motor-distance $l$. 
This indicates that the motor-distance is an appropriate scaling for the drones' flow geometry.

In between the two jets, a region of reduced flow speed can be observed. 
This depression is due to the separation of the drones' rotors and the blocking of the flow by its main body. 
Such flow structures have previously been observed in CFD simulations~\cite{Crazzolara2019} and validation studies~\cite{Villa2016}, and are consistent with our results. 
Afterwards, the jets start to converge at about one length scale below the rotor plane. 
From about 2.5 length scales below the drone, the jets appear to have fully merged.

\begin{figure}[t!]
    \centering
    \includegraphics{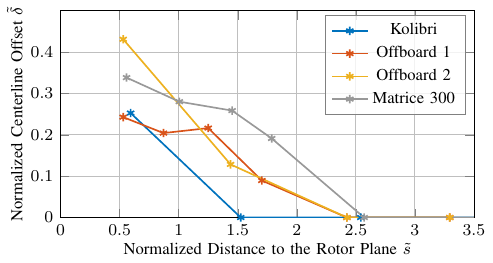}
    \vspace*{-7pt}
    \caption{Distance $\tilde{\delta}$ of the flow tube centerline to the $\mathbf{z}_\bfr$ axis. 
    Close to the rotor-plane of the vehicle (left side) the flow has not fully developed. From 2 to 3  length scales below the drone the flow fully develops with the highest velocity being measured on the negative z-axis of the drone.}
    \label{fig:nearfield_angle}
    \vspace*{-12pt}
\end{figure}

Figure~\ref{fig:nearfield_angle} quantitatively describes the merging of the individual jets. 
The distance of the maximum propeller flow velocity centerline to the drones' negative $\mathbf{z}_\bfr$ axis is considered. 
This distance is estimated by first extracting the $\tilde{y}$-axis velocity profile for different distances below the drone.
Then this empirical profile is approximated with a $z_\bfr$-axis symmetric bimodal Gaussian where the means are shifted $\pm\tilde\delta$ along the $\tilde{y}$-axis. 
In agreement with related works~\cite{Crazzolara2019} we find that for our drones the individual rotor flows merge between 1.5 and 2.5 motor-distances below the drone.

\vspace*{-4pt}
\section{Results: Far Field}

From the preliminary considerations of the airflow close to the vehicle, we see that the individual contributions from the quadrotor's four propellers merge about 2.5-length scales below the drone. 
In this section, we present our novel experimental findings to answer the question: `How much wind do the quadrotors make away from their propellers in the far field?'.
For the remainder of this section, cylindrical coordinates are used. 
The data is binned radially along $r$ and analyzed for downstream coordinates $s$, following the convention introduced in Fig.~\ref{fig:quadrotor}.

\begin{figure}[t]
    \centering
    \includegraphics{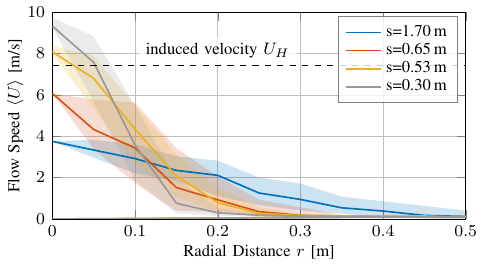}
    \vspace*{-9pt}
    \caption{Radial velocity profiles for the \emph{Kolibri} drone. The colors represent different height levels. 
    The shaded area indicates the $1\sigma$ uncertainty region around the mean, the line shows the median value.}
    \label{fig:farfield}
    \vspace*{-4pt}
\end{figure}

\begin{figure}[t]
    \centering
    \includegraphics{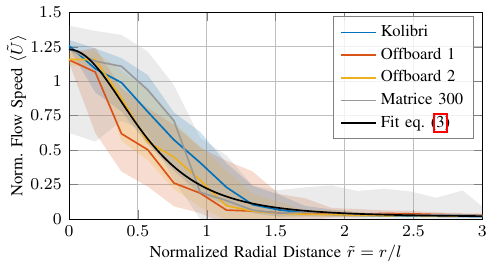}
    \vspace*{-8pt}
    \caption{Radial velocity profiles for the drones with a planar propeller layout at a normalized height of $\tilde{s}=3.0$. 
    The shaded area indicates  $1\sigma$ uncertainty around the mean, the line shows the median value. 
    The proposed normalization leads to all curves being similar in measurement accuracy. 
    The black line represents the radial velocity profile for a turbulent jet eq.~\eqref{eq:radialprofile}.}
    \label{fig:normalized_farfield}
    \vspace*{-15pt}
\end{figure}

\begin{figure*}
    \centering
    \includegraphics{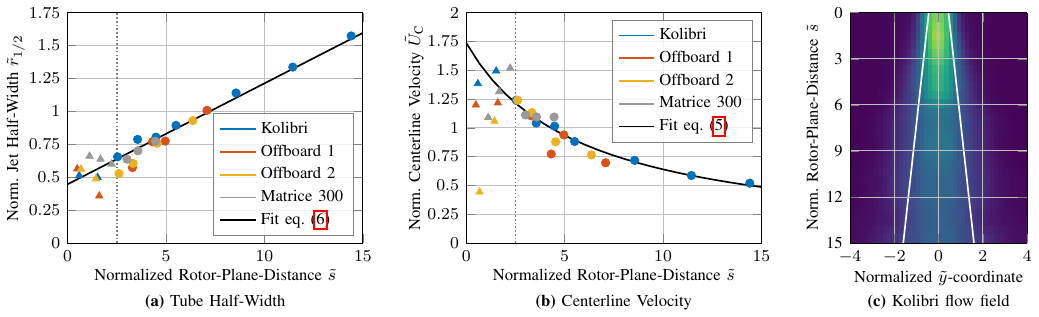}
    \vspace*{-8pt}
    \caption{Downstream scaling parameters of the flow. Only data points in the far field region (marked with a circle) are used to fit the function predicted by turbulent jet theory. {\textbf{(a)} The} plot depicts the (normalized) tube half-width \raisebox{1pt}{\scalebox{0.9}{$\tilde{r}_{1/2}$}} as a function of the (normalized) distance to the rotor plane $\tilde{s}$. We recover a characteristic spreading angle of \unit[10]{deg}. {\textbf{(b)} The} plot focuses on the (normalized) centerline velocity $\tilde{U}_\text{C}$ as a function of the (normalized) distance to the rotor plane. We recover the characteristic inverse $Bd/(s-s_0)$ scaling from eq.~\eqref{eq:centerline}. The parameters for \eqref{eq:centerline} and \eqref{eq:spread} are fitted jointly to the data. {\textbf{(c)} A} visualization of the measured \emph{Kolibri} drone flow field. The tube half-width from (a) is overlaid in white.
    }
    \label{fig:expansion}
    \vspace*{-15pt}
\end{figure*}

\vspace*{-7pt}
\subsection{Radial Velocity Profiles} 
\label{sec:radial_profiles}
Figure~\ref{fig:farfield} exemplarily shows radial profiles of the mean velocity for the \emph{Kolibri} drone at different heights. 
First, the measured peak velocities exceed the calculated induced velocity at hover by \unit[25]{\%} at maximum. 
This mismatch with the calculated propeller-induced flow is expected as the velocity profile in the merged flow is not constant across the cross-section and by conservation of momentum the peak velocity is higher. 
Second, we compare measurements at different $s$-coordinates and observe that the mean flow velocities all decay radially from the drone. 
Third, the downstream peak velocity rapidly decreases away from the drone, while the flow domain spreads out.

We now look at the data for multiple drones using the normalization introduced in Sec.~\ref{sec:normalization}. 
The plot in Fig.~\ref{fig:normalized_farfield} corresponds to a normalized $\tilde{s}$ level at 3 length scales (motor-distance) below the drone. 
The normalized, time-averaged measured velocity $\langle \tilde{U}\rangle$ is plotted as a function of the normalized radial coordinate $\tilde{r}$ and each color now corresponds to one drone. 
We can clearly see that, within the given measurement accuracy, the curves overlap which empirically validates the proposed normalization as it makes the airflow around different vehicles similar. 
Additionally, the radial velocity profile across all drones is well captured fitting eq.\eqref{eq:radialprofile}, inset in black in Fig.~\ref{fig:normalized_farfield}. 
This provides evidence that the combined rotor flow can be treated as a turbulent jet.

\vspace*{-7pt}
\subsection{Jet Flow Scaling}
\label{sec:scaling}

Next, we focus on the expansion of a turbulent jet which is described by eq.~\eqref{eq:spread}. 
For a turbulent jet, the flow expands linearly as a cone with approximately \unit[12]{deg} opening angle~\cite{Pope2000}. %
Furthermore, the jet's centerline velocity scales inversely proportional to the distance as described by eq.~\eqref{eq:centerline}.

In Figure~\ref{fig:expansion} (a) and (b) we find good agreement with these two key characteristics for the combined rotor flow of a quadrotor after $\tilde{s} = 2.5$. 
Figure~\ref{fig:expansion}a and b show the normalized half-width $\tilde{r}_{1/2}$ and the normalized centerline velocity $\langle \tilde{U}_\text{C}\rangle$ of the radial velocity profiles as a function of the normalized distance $\tilde{s}$ to the drone's rotor plane. 
The half-width and velocity have been obtained by fitting the velocity profile eq.~\eqref{eq:radialprofile} to the measurements at each $\tilde{s}$-slice for each drone. 
The triangles indicate datapoints located in the near-field (not considered for the fit) and circles indicate measurements from the far field (considered for the fit). The black lines represent a fit using the function form prescribed by the turbulent jet model eq.~\eqref{eq:centerline} and \eqref{eq:spread}. 
From the fit we obtain \vspace*{-2pt}
\begin{equation}
    \begin{aligned}
        B\,d = 10.11\,, \quad S = 0.07668\,\text{~and}\quad s_0 = -5.817\; 
    \end{aligned}
    \vspace*{-2pt}
    \label{eq:params}
\end{equation} 
for the unified flow.
Note that only drones with uncanted propeller configurations are used to obtain the fit. 
Figure~\ref{fig:expansion}c shows visualization of the \emph{Kolibri} far-field velocity profile with the white lines indicating the half-width (see Fig.~\ref{fig:expansion}a). 

\begin{figure}[t]
    \centering
    \vspace*{6pt}
    \includegraphics{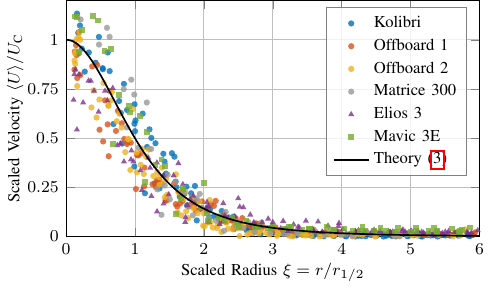}
    \vspace*{-12pt}
    \caption{Unified formulation of the flow. 
    The measurement samples from all planar drones across the entire far field show good agreement with turbulent jet theory eq.~\eqref{eq:radialprofile} (black line). 
    Drones with canted propellers show a slightly narrower or a wider jet, depending on the geometry.
    }
    \label{fig:alldata}
    \vspace*{-16pt}
\end{figure}

The scaling parameters~\eqref{eq:params} can be used to fully describe the turbulent jet flow in eq.~\eqref{eq:radialprofile}. This velocity profile from eq.~\eqref{eq:radialprofile} is shown with the black line in Fig.~\ref{fig:alldata} and is not fitted to the data. The measurement samples are taken across the entire far field and measurement scaling is computed using eq.~\eqref{eq:centerline} and~\eqref{eq:spread} with the parameters above. The good agreement with the theory shows that the measured velocity of a quadrotor field in both axial and radial dimension is well captured by the theory of turbulent jets.

\begin{table}[b]
    \centering
    \vspace*{-16pt}
    \caption{Hypothesis Test ($\alpha = 0.05$):~~ $H_0: \mu(\epsilon)=0$,~~ $H_A: \mu(\epsilon)>0$ }
    \label{tab:distribution}
    \vspace*{-3pt}
    \setlength{\tabcolsep}{4pt}
    \begin{tabularx}{1\linewidth}{l|XXXXXX}
    Interval \hspace{1.6cm}$ \xi \in $ &  [0~ 1) & [1~2) & [2~ 3) & [3~ 4) & [4~ 5) & [5~ 6)\\
    \midrule
    Reject $H_0$, e.g., $\mu(\epsilon) > 0 $  & False & False & True & True & True & False
    \end{tabularx}
\end{table}

We note that the measured velocities in Fig.~\ref{fig:alldata} for uncanted drone designs are well-centered around the theoretical curve in the region for low $\xi$ but seem to fall below the theory curve $U_\text{theory}$ eq.~\eqref{eq:radialprofile} around ${\xi > 2}$ until the measurement noise dominates from ${\xi > 5}$ onwards. To corroborate this visual impression, a one-sided $t$-test for the mean of the distribution of measurement residuals ${\epsilon = U_\text{theory} - \langle U \rangle / U_C}$ is conducted. We verified the normality of the residuals with a $\chi^2$ goodness-of-fit test. The $t$-test results (see Tab.~\ref{tab:distribution}) show that for ${\xi > 2}$ the theory overpredicts the velocity. This is a well-known result for turbulent jets~\cite{Pope2000}. However, in practice it can be safely ignored due to the negligible magnitude of the mismatch w.r.t typical ambient airflow.

In Fig.~\ref{fig:alldata} the two drones with canted propeller planes are also included in the plot and represented with squares and triangles. The \emph{Mavic 3E} design features an inward cant, effectively focusing the flow. This is consistent with the observation that the jet is narrower with slightly higher centerline velocities. The \emph{Elios 3} on the other hand has outwardly canted propellers which leads to a wider jet with lower centerline velocities but a larger half-width. This effect is especially notable around ${\xi = 0}$ and ${\xi = 2.5}$.

\vspace*{-6pt}
\subsection{Unified Model}
To use the unified model describing the far field flow of a quadrotor, we combine the results demonstrating normalization and scaling. As  demonstrated in Fig.~\ref{fig:normalized_farfield}, the normalization of the drones' propeller-induced flow and geometry of Sec.~\ref{sec:normalization} makes drones of different  mass and size similar. The recovered scaling parameters for eq.~\eqref{eq:spread} and~\eqref{eq:centerline} enable us to leverage self-similarity to fully describe the downwash of the quadrotor as a turbulent jet (see Fig.~\ref{fig:alldata}).

To calculate the time-averaged velocity $\langle U\rangle$ in the far field below a quadrotor at a point $\mathbf{p} = [s~~r~~\theta]^\top$, perform the following operations:
\begin{enumerate}
    \item Normalization parameter: Use 
    the physical parameters (mass, propeller size, number of propellers) of the quadrotor to calculate the induced velocity at hover $U_{\rm{H}}$ using eq.~\eqref{eq:v_ind_h} and the motor distance $l$. 
    \item Normalize the point's length-scale: $\tilde{s} = s/l$, $\tilde{r} = r/l$.
    \item Calculate the normalized turbulent jet scaling $\tilde{r}_{1/2}$ and $\tilde{U}_\text{C}$ for the normalized point. This is done with eq.~\eqref{eq:centerline} (set {$U_\text{J} = 1$}) and~\eqref{eq:spread} together with the parameters \eqref{eq:params}.
    \item Calculate the scaled radial position {$\xi = \tilde{r}/\tilde{r}_{1/2}$}.
    \item Scale the centerline velocity {$U_\text{C} = \tilde{U}_\text{C} \cdot U_{\rm{H}}$}.
    \item Finally, use $\xi$ and $U_\text{C}$ to  evaluate the time-averaged flow speed using the turbulent jet equation~\eqref{eq:radialprofile}.
\end{enumerate}
Note that, in case of interest, the entrained radial mean flow component $\langle V \rangle$ can be obtained through continuity \cite{Pope2000}.

\section{Controller Integration}
To highlight that our turbulent-jet model is relevant to real-world multi-agent scenarios, we conclude our work by demonstrating how the model can be used to compute a feed-forward downwash compensation in a scenario where one drone needs to pass below another drone.

\vspace*{-6pt}
\subsection{Downwash Compensation}
When flying  below another drone, the lower drone needs to additional power to maintain hover thrust.
Similar to eq.~\eqref{eq:v_ind_h}, momentum theory can be used to calculate the airspeed $U_H'$ below the propeller.
However, in this case the air above the propeller is not at rest but moving downwards with $U_D$~\cite{1995:Prouty},
\begin{equation}
    U_H' = \frac{U_D}{2} + \sqrt{\left(\frac{U_D}{2}\right)^2 + U^2_H}\;,
\end{equation}
where $U_H$ from is the induced velocity at hover in still air from eq.~\eqref{eq:v_ind_h}, and $U_D$ is calculated according to the unified model presented in the previous section. \rebuttal{Both $U_D$ and $U'_H$ depend on the relative position $\Delta \mathbf p$ of the drones.}
The total aerodynamic power required to hover inside the downwash is ${P_H' = U_H' \cdot T_H}$. 
Introducing the relative induced velocity ${\alpha = U_D/U_H}$ we get the relative power $\beta$ as
\begin{equation}
    \beta := \frac{P_H'}{P_H} = \frac{\alpha}{2} + \frac{\sqrt{\alpha^2 + 4}}{2}\;.
    \label{eq:powercorrection}
\end{equation}

The electronic speed controllers (ESC) on the \emph{Kolibri} drone exhibit a quadratic relationship between the throttle command and the power consumption. 
Consequently, the throttle setpoint computed by the controller is scaled with $\sqrt{\beta}$ before sending it to the vehicle's motor controllers. \rebuttal{This scaling factor is time-varying as it depends on $\Delta \mathbf p$.}
We implement the downwash compensation on top of an MPC controller~\cite{foehn2022agilicious} by scaling the throttle command according to~\eqref{eq:powercorrection}.

\vspace*{-6pt}
\subsection{Experimental Results}
\begin{figure}[t]
    \centering
    \includegraphics{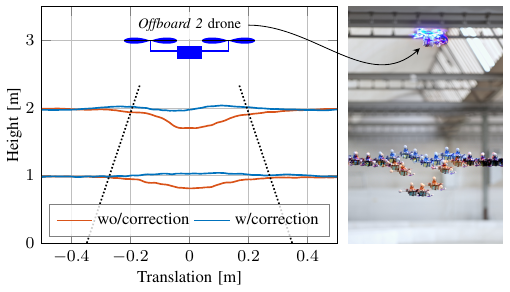}
    \vspace*{-6pt}
    \caption{Experiment showing that our downwash model can be used to improve the control performance in a multi-drone scenario. 
    The dotted line \rebuttal{starting at 2.5 length scales below \emph{Offboard 2}} indicates the half-width calculated from eq.~\eqref{eq:spread}. \rebuttal{The depicted drone is true-to-scale in width.}
    The right side shows a time-lapse photograph of the run at \unit[2]{m}.}
    \label{fig:controller}
    \vspace*{-6pt}
\end{figure}

To demonstrate the effectiveness of the control scheme, we perform the following experiment: the \emph{Offboard 2} drones hovers at \unit[3]{m} height and the much smaller \emph{Kolibri} passes 1 and 2 meters exactly below at a translation speed of 1 length-scale per second. 
The results are shown in Fig.~\ref{fig:controller} and Table~\ref{tab:controller} and one can see that the baseline controller without the proposed downwash correction has a much higher altitude tracking error compared to the corrected controller. Directly below the drone the controller without correction deviates up to \unit[30]{cm} from the reference height.
Comparing the \rebuttal{RMSE in height} we see a 3 to 5-fold improvement with the proposed controller \rebuttal{and observe that the performance improvement is larger in stronger downwash (e.g., closer to the upper drone).}

\begin{table}[t]
    \centering
    \color{black}
    \caption{Height Tracking Performance of the Controller}
    \label{tab:controller}
    \vspace*{-6pt}
    \begin{tabularx}{1\linewidth}{Xc|rr}
    \toprule
    Controller & Height [m] & RMSE [mm] & Mean Err. [mm] \\ \midrule
    Without Correction & \unit[1]{m} & 69.1 & -37.6\\
    With Correction & \unit[1]{m} & 24.8 & 18.9 \\
    \midrule
    Without Correction & \unit[2]{m} & 134.2 & -87.1 \\
    With Correction & \unit[2]{m} & 25.4 & 8.1 \\
    \bottomrule
    \end{tabularx}
    \vspace*{-12pt}
\end{table}

\section{Conclusion and Discussion}
This work presented a model for the far field flow of a drone, based on classical turbulent jet theory. 
We recorded a large-scale dataset comprising six drones and containing over \unit[16]{h} of flight data in a motion-capture system while measuring the airflow with a hot-ball flow probe. 
We studied the flow in the near-field flow close to the drone and found that the individual flows from the propellers merge approximately 2.5 motor-distance length scales below the drone. 
In this near-flow region, the interactions between the flow and the drone are dominant which can only be fully captured through CFD simulations. 
However, when considering the effects of the induced flow on the environment and other agents, the primary concern is the far-field extending from about two length scales downwards. 

Through the use of \rebuttal{appropriate scaling laws for turbulent flows}, we developed a unified model that describes the far field flow below the drone \rebuttal{as a turbulent jet}.
The experiments show that, despite the simplicity of the model, it accurately describes the flow for a wide range of drones \rebuttal{at hover}. 
\rebuttal{Jet flows are well-established in fluid mechanics.
For example, in the case of cross-flows \cite{Manesh2013}, jets include empirical scalings for the center-line shape, relevant to vertical and forward flight~\cite{1995:Prouty}, and rotor induced-vortex flows \cite{Gardner2019}. %
Moreover, experiments on impinging jets~\cite{Barata1993} and similarity-solutions for wall-bounded jet flows~\cite{George2000} help understand modulations in ground effect and near boundaries as a first step towards confined environments with internal circulation. 
Showing that the induced flow of a drone is a turbulent jet, we believe that applying the above results from fluid mechanics %
presents an interesting avenue for further research.}\!

Finally, we demonstrated that the accuracy and efficiency make our model ideally suited for integration in a controller for multi-agent scenarios. 
We believe that this is an important step towards safer and less intrusive drones, an aspect becoming increasingly important with the increasing adoption of quadrotors. 
Understanding the flow also enables optimized sensor placement for scientific applications and potentially higher efficiency when quadrotors are deployed in an agricultural context.

{\small
\balance
\bibliographystyle{IEEEtran}
\bibliography{references}
}

\end{document}